\documentclass[information,article,accept,moreauthors,pdftex]{Definitions/mdpi} 

\firstpage{1} 
\pubvolume{1}
\issuenum{1}
\articlenumber{0}
\pubyear{2022}
\copyrightyear{2022}
\externaleditor{Academic Editor: Firstname Lastname}
\datereceived{} 
\dateaccepted{} 
\datepublished{} 
\hreflink{https://dx.doi.org/} 

\usepackage{multirow}

\Title{A Literature Survey of Recent Advances in Chatbots}

\TitleCitation{A Literature Survey of Recent Advances in Chatbots}

\Author{Guendalina Caldarini 
 *$^{\dagger,\ddagger}$\orcidB{}{}, Sardar Jaf $^{\ddagger}$\orcidA{} and Kenneth McGarry \orcidC{}}

\AuthorNames{Guendalina Caldarini, Sardar Jaf and Kenneth McGarry}

\AuthorCitation{Caldarini, G.; Jaf, S.; McGarry, K.}

\address[1]{%
The University 
 of Sunderland, Faculty of Technology, Sunderland, SR1 3SD, UK
 
sardar.jaf@sunderland.ac.uk (S.J.); guendalina.caldarini@research.sunderland.ac.uk (G.C.);
 ken.mcgarry@sunderland.ac.uk 
 (K.M.)}

\corres{\hangafter=1 \hangindent=1.05em \hspace{-0.82em} Correspondence: guendalina.caldarini@research.sunderland.ac.uk}

\firstnote{\hangafter=1 \hangindent=1.05em \hspace{-0.82em} Current address: The University of Sunderland, Sunderland, SR1 3SD, UK
.} 

\secondnote{\hangafter=1 \hangindent=1.05em \hspace{-0.82em} These authors contributed equally to this work.}

\abstract{Chatbots are intelligent conversational computer systems designed to mimic human conversation to enable automated online guidance and support. The increased benefits of chatbots led to their wide adoption by many industries in order to provide virtual assistance to customers. Chatbots utilise methods and algorithms from two Artificial Intelligence domains: Natural Language Processing and Machine Learning. However, there are many challenges and limitations in their application. In this survey we review recent advances on chatbots, where Artificial Intelligence and Natural Language processing are used. We highlight the main challenges and limitations of current work and make recommendations for future research investigation}


\begin{document}

\section{Introduction}
\label{sec:introduction}

Chatbots are intelligent conversational computer programs that mimic human conversation in its natural form~\cite{jia_study_2003,sojasingarayar_seq2seq_2020,bala_chat-bot_2017}. A~chatbot can process user input and produce an output~\cite{ayanouz_smart_2020,kumar_review_2020}. Usually, chatbots take natural language text as input, and~the output should be the most relevant output to the user input sentence. Chatbots can also be defined as ``online human-computer dialogue system(s) with natural language''~\cite{cahn_chatbot_2017}. Chatbots constitute therefore an automated dialogue system, that can attend to thousands of potential users at~once.

Chatbots are currently applied to a variety of different fields and applications, spanning from education to e-commerce, encompassing healthcare and entertainment. Therefore, chatbots can provide both support in different fields as well as entertainment to users~\cite{okuda_ai-based_2018}; this is the case for chatbots such as Mitsuku and Jessie Humani, ``small talk'' oriented chatbots that could provide a sense of social connection~\cite{kompatsiaris_why_2017}. Chatbots appear, in~fact, to~be more engaging to the user than the static Frequently Asked Questions (FAQ) page of a website. At~the same time, chatbots can simultaneously assist multiple users, thus resulting more productive and less expensive compared to human customer supports services. In~addition to support and assistance to customers, chatbots can be used for providing entertainment and companionship for the end user~\cite{costa_conversing_2018}. Nonetheless, different levels of embodiment-the way chatbots are human-like~\cite{go_humanizing_2019}-and disclosure-how and when the nature of the chatbot is revealed to the user-seem to impact users' engagement with and trust in chatbots~\cite{luo_frontiers_2019}).

In recent years, with~the commoditization and the increase of computational power and the sharing of open source technologies and frameworks, chatbots programmes have become increasingly common. Recent developments in Artificial Intelligence and Natural Language Processing techniques have made chatbots easier to implement, more flexible in terms of application and maintainability, and~increasingly capable to mimic human conversation. However, human-chatbot interaction is not perfect; some areas for improvements are contextual and emotional understanding and gender biases. Chatbots are, in~fact, less able to understand conversational context~\cite{christensen_context-aware_2018} and emotional linguistic cues compared to humans, which affects their ability to converse in a more entertaining and friendly manner~\cite{fernandes_nlp_2018}. At~the same time, chatbots tend to take on traditionally feminine roles which they perform with traditionally feminine features and often displaying stereotypical behaviour, revealing a gender bias in chatbots' implementation and application~\cite{costa_conversing_2018}.

Since chatbots are so widespread and applied to many different fields, improvements in their implementations and evaluation constitute important research topics. The~main contributions of this paper are: (i) extensive survey of the literature work on chatbots as well as the state of the art on chatbots' implementation methods, with~a focus on Deep Learning algorithms, (ii) the identification of the challenges and limitations of chatbots implementation and application, and~(iii) recommendation for future research on~chatbot.

The rest of this article is organized as follows: Section~\ref{sec:background} provides some background on chatbots and their evolution through time, Section~\ref{sec:methodology} describes the methodology, Section~\ref{sec:literature} presents an analysis of the state of the art in terms of chatbots Deep Learning algorithms; including the datasets used for training and evaluation methods, in~Section~\ref{sec:related-work} we will discuss related works and we conclude the paper in Section~\ref{sec:conclusion}.

\section{Chatbots~Background}
\label{sec:background}
Although the quest for being able to create something that can understand and communicate with its creator has deep roots in human history, Alan Turing is thought to be the first person to have conceptualised the idea of a chatbot in 1950, when he proposed the question: ``Can machines think?''. Turing's description of the behaviour of an intelligent machine evokes the commonly understood concept of a chatbot~\cite{turing_icomputing_1950}.

Chatbots have evolved with the progressive increase in computational capabilities and advances in Natural Language Processing tools and techniques. The~first implementation of a chatbot, which relied heavily on linguistic rules and pattern matching techniques, was achieved in 1966 with the development of ELIZA. It could communicate with the user through keyword matching program. It searches for an appropriate transformation rule to reformulate the input and provide an output, i.e.,~an answer to the user. Eliza was a landmark system that stimulated further research in the field. Nonetheless, ELIZA's scope of knowledge was limited because it depended on minimal context identification and, generally, pattern matching rules are not flexible to be easily implemented in new domains~\cite{weizenbaum_eliza--computer_1966,shum_eliza_2018,zemcik_brief_2019}. 

A marked evolution in chatbot in the 1980s is the use of Artificial Intelligent. A.L.I.C.E. (Artificial Intelligent Internet Computer Entity) is based on the Artificial Intelligence Mark-up Language (AIML), which is an extension of XML. It was developed especially so that dialogue pattern knowledge could be added to A.L.I.C.E.'s software to expand its knowledge base. Data objects in AIML are composed of topics and categories. Categories are the basic unit of knowledge, which are comprised of a rule to match user inputs to chatbot's outputs. The~user input is represented by rule patterns, while the chatbot's output is defined by rule template, A.L.I.C.E. knowledge base. The~addition of new data objects in AIML represented a significant improvement on previous pattern matching systems since the knowledgebase was easily expandable. Furthermore, ChatScript, the~successor of AIML, was also the base technology behind other Loebner's prize-winning chatbots. The~main idea behind this innovative technology was to match textual inputs from users to a topic, and~each topic would have specific rule associated with it to generate an output. ChatScript ushered in a new era for chatbots' technology evolution. It started shifting the focus towards semantic analysis and understanding~\cite{bradesko_survey_2012,wilcox_winning_2014,abushawar_alice_2015,cahn_chatbot_2017,shum_eliza_2018,zemcik_brief_2019}.

The main limitation in relying on rules and pattern matching in chatbots is they are domain dependent, which makes them inflexible as they rely on manually written rules for specific domains. With~the recent advances in machine learning techniques and Natural Language Processing tools combined with the availability of computational power, new frameworks and algorithms were created to implement ``advanced'' chatbots without relying on rules and pattern matching techniques and encouraged the commercial use of chatbots. The~application of machine learning algorithms in chatbots has been investigated and new architectures of chatbots have~emerged.

The application of chatbots has expanded with the emergence of Deep Learning algorithms. One of the new, and~the most interesting application, is the development of smart personal assistants (such as Amazon's Alexa, Apple's Siri, Google's Google Assistant, Microsoft's Cortana, and~IBM's Watson). Personal assistants chatbots or conversational agents that can usually communicate with the user through voice are usually integrated in smartphones, smartwatches, dedicated home speakers and monitors, and~even cars. For~example, when the user utters a wake word or phrase the device activates, and~the smart personal assistant starts to listen. Through Natural Language Understanding the assistant can then understand commands and answer the user's requests, usually by providing pieces of information (e.g., ``Alexa, what's the weather today in Los Angeles?`` In Los Angeles the weather is sunny and there are $75^\circ F$''), or~by completing tasks (e.g., ``Ok Google, play my morning playlist on Spotify''). Nonetheless, the~task of understanding human language has proven to be quite challenging because of tonal, regional, local, and~even personal variations in human~speech.

All smart personal assistants present the same core characteristics in terms of technologies used, user interface and functionalities. Some chatbots have, however, a~more developed personality than others, and~the most developed ones can also provide entertainment and not merely assistance with day-to-day tasks; these chatbots are referred to as social chatbots. An~interesting example of a social chatbot is Microsoft's XiaoIce. XiaoIce is meant to be a long-term companion to the user, and~in order to achieve high user engagement it has been designed to have a personality, an~Intelligent Quotient (IQ) and an Emotional Quotient (EQ). Knowledge and memory modelling, image and natural language comprehension, reasoning, generation, and~prediction are all examples of IQ capabilities. These are critical components of the development of dialogue abilities. They are required for social chatbots to meet users' specific needs and assist them. The~most critical and sophisticated ability is Core Chat, which can engage in lengthy and open-domain conversations with users. Empathy and social skills are two critical components of EQ. The~conversational engine of XiaoIce uses a dialogue manager to keep track of the state of the conversation and selects either the Core Chat (the open domain Generative component) or the dialogue skill in order to generate a response. Therefore, the~model incorporates both Information-Retrieval and Generative capabilities~\cite{dormehl_microsofts_2018,spencer_much_2018,zhou_design_2019}.

\section{Methodology}
\label{sec:methodology}

Our approach for conducting this literature survey study consists of two stages. Each stage involves several activities. In~the first stage, we identify relevant search terms to literature work on the topic, and~then we identify appropriate databases of research articles. Then, we collect research articles on chatbots from the selected databases. These activities are focused on information gathering about the topic. The~second stage of our study involves the analysis of the retrieved articles. We focus on classifying the articles to different groups based on four aspects of chatbots: design, implementation, application and evaluation methods reported in the literature. In~the following subsections, we give details of those~activities.

\subsection{Stage One: Information~Gathering}
\label{subsec:information-gathering}
\unskip

\subsubsection*{Search Terms and Databases~Identification}
We have used three large publishers' databases for identifying research articles on chatbots. These are IEEE, ScienceDirect and Springer. These databases provide a good assortment of peer reviewed articles in the fields of Natural Language Processing, Artificial Intelligence, and~Human-Computer Interaction. In~addition to those databases, we have searched for publication on  arXiv, Google Scholar and JSTOR because they provided a considerable number of publications and material. We selected publications from various fields, including information technology; computer sciences; computer engineering; communication and social arts.
All the databases, indexes and repositories we selected to retrieve articles from them were because we had free access to the publications, and~they allowed us to query articles through search words. The~Scopus database, for~example, provided valuable information regarding chatbot related publications-as seen in Figure~\ref{fig:1}-but unfortunately, it did not allow us to search articles by keyword.

We initially started querying the selected databases using a seed search term ``chatbot'' which broadly described the subject matter. From~some of the articles identified using our seed term, we identified the following new search terms: ``conversational modelling'', ``conversation systems'', ``conversational system'', ``conversational entities'', ``conversational agents'', ``embodied conversational agents'' and ``Human-Computer Conversational Systems''.~Ref. \citep{radziwill_evaluating_2019}, for~example, define chatbots as a subcategory of conversational agents that are not usually embodied. Some of these keywords were also mentioned in relevant works of literature: ``conversational entities'' appears in~\cite{sheehan_customer_2020}; ``conversational agents'' appears in~\cite{adamopoulou_overview_2020,nuruzzaman_survey_2018}, and~in~\cite{ketakee_chatbots_2017}; ``conversation system'' appears in the title of~\cite{yan_learning_2016,abdul-kader_survey_2015}.

\subsection{Stage Two: Article Filtering and~Reviewing}
\unskip
\subsubsection{Filtering~Articles}
\begin{figure}[H]
\includegraphics[width=12cm]{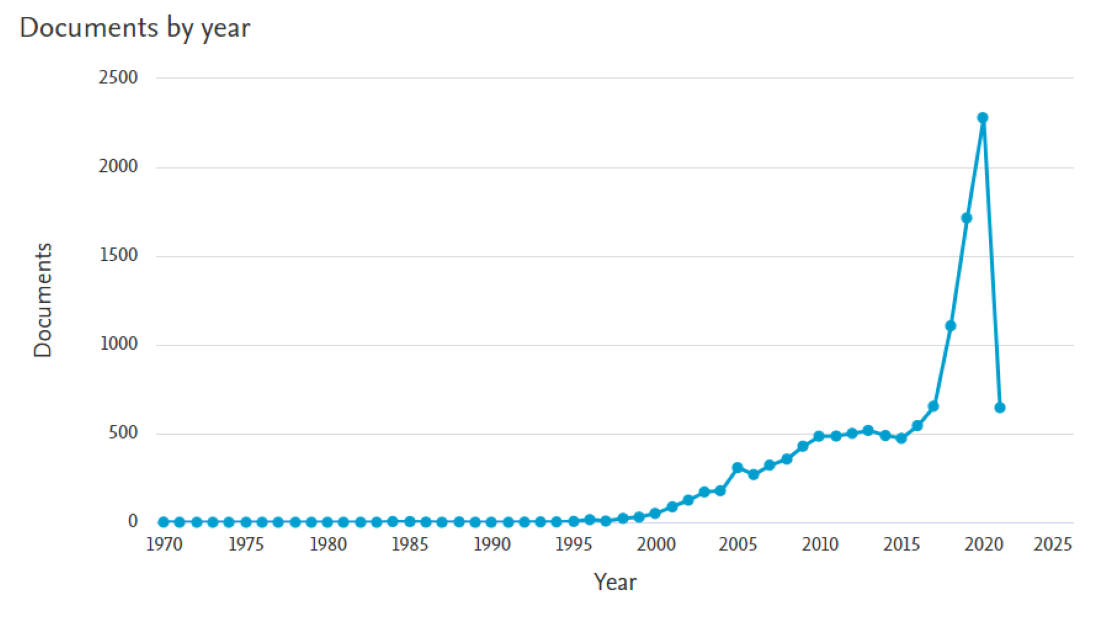}
\caption{Search Results 
 from Scopus, from~1970 to 2021 for the keywords ``chatbot'' or ``conversational agents'' or ``conversation~system''.\label{fig:1}}
\end{figure}

\textls[-10]{After we have identified several search terms, we have queried the databases. The~search} terms returned high quantities of research articles. The~initial search result returned thousands of pieces of literature spanning from 1970s to 2021. In~order to reduce the number of articles for analysis to a manageable number, we have filtered the search results based on publication date. We focused on articles published between 2007 and 2021. Our aim was to focus on articles in a specific year range and with relevant title to our intended study.  Based on the analysis from Scopus database, there has been increased publication on chatbot from the year of 2000. Figure~\ref{fig:1} shows a steady increase in research publication on chatbot with a small dip in 2005 but a rapid peak occurring in 2020.  For~this study we focused on articles published from 2007 because this was the start of publication peak on chatbot. This has resulted in a total of 62,254 articles published in the selected databases and other publication repositories between 2007 and 2021. In~Table~\ref{tab:1} we present the number of articles retrieved using our set of search terms in different databases. It is worth noting that the search term ``human-computer conversational systems'' attracted no articles in several databases. ArXiv or Google Scholar databases returned more recently published articles.

After the first filtering step, we applied a second filtering operation by selecting articles based on their title. The~objective was to focus only on relevant articles for our~study.

Initially, we searched for articles in journal databases, such as IEEE; ScienceDirect; Springer; and JSTOR as we had institutional access to them to retrieve full text of articles, with~specific keywords. The~search result contained several thousands of peer reviewed published articles. We analyzed the title of each article returned to determine its relevance to our study. We selected only the relevant articles that provided concrete examples of chatbots' technical implementations and development. Additionally, we considered publications that focused on literature review and literature survey of chatbots. This approach enabled us to get a picture of the state of the art and the evolution of the field over time--Section{~\ref{sec:background}} presents the evolution of cahtbots over a period of time. To~maintain the original focus of our study, we discarded articles that focused on the marketing advantages of chatbots or social studies involving chatbots. We also discarded articles that referenced chatbots as ``means to an end'' rather than the final product. Therefore, we decided to discard these kinds of articles because our aim was to survey Deep Learning algorithms and Natural Language Processing techniques used in the latest chatbots' implementations. At~the end of this step, we selected 59 articles for our~study.

The second step in collecting articles for reviewing involved searching arXiv (a repository of preprints, available at 
 \url{https://arxiv.org/}, access date: 06/01/2022), where we have used a set of search word/phrases to identify additional articles. We followed the same filtering process as that applied to the journal databases. To~avoid any overlap between the articles selected from the main databases (IEEE, ScienceDirect, Springer, and~JSTOR) and arXiv, we used a reference management software (Zotero (\url{https://www.zotero.org/})), which can be utilized to identify if an article had been selected twice. This way we avoided overlap between the different databases and arXiv repository. By~the end of this step, we selected 56 articles for our study. We combined these articles to the previous 56 articles, which we obtained by searching several journal databases, and~obtained a total of 115~articles.

Finally, we studied the bibliographies of the 115 articles to identify more articles that seemed pertinent. We used Google scholar to retrieve the full text of potential articles that appeared in the bibliography of 115 articles. This process allowed us to obtain a further 201 relevant articles for our study. Thus, the~total number of articles at our disposal for reviewing was 316 articles.

 Table{~\ref{tab:1}} presents the search result returned from various sources we have used in this study to retrieve articles published on chatbot between 2007 and 2021. Several journal databases, repositories and search engines are used, where we searched for articles by search terms related to chatbots. It is worth noting that we used Google scholar to search for articles by title instead of search terms. However, the~fifth row in the table displays the number of publications for each search term. Because~the search engine returned a substantial number of articles (over 45k articles) we did not process them. The~201 selected articles in the fifth row are retrieved from Google scholar by searching for those article title identified in the bibliography of some previously identified articles (the 115 articles that were retrieved from several journal databases and arXiv).

\begin{table}[H] 
\small
\caption{Number of Article by  search~terms.\label{tab:1}}

\begin{adjustwidth}{-\extralength}{0cm}
\centering 
\begin{tabular}{m{26mm} c m{26mm} m{26mm} m{27.4mm}}
\toprule
\textbf{Database and Repositories }	& \textbf{Keyword}	& \textbf{Total Number of Articles} & \textbf{Total Articles between 2007 and 2021} & \textbf{Number of Articles Selected for Reviewing } \\
\midrule
\multirow{8}{*}{IEEE}	& chatbot	   & 666     & 664	   & \multirow{8}{*}{22}\\
 & conversational modelling	   & 1152     & 831	   & \\ 
& conversational system	& 42 & 23 \\
& conversation system	& 3099	& 2321 \\
& conversational entities & 	51	& 42 \\
& conversational agents	& 590	& 503 \\
& embodied conversational agents & 	160	&  137 \\
& human-computer conversational systems	& 217	&  181 \\
\midrule
\multirow{8}{*}{ScienceDirect}	& chatbot	   & 1063     & 1058	   & \multirow{8}{*}{20}\\
& conversational modelling	& 188	& 105 \\
& conversational system	& 318	& 119\\
& conversation system	& 185	& 137\\
& conversational entities	& 9 & 	8\\
& conversational agents	& 674	& 597\\
& embodied conversational agents & 	282	& 243\\
& human-computer conversational systems	& 2	& 2\\
\midrule
\multirow{8}{*}{Springer}	& chatbot	   & 2046   & 2010	   & \multirow{8}{*}{16}\\
& conversational modelling	& 441	& 293 \\
& conversational system & 	862	& 564\\
& conversation system	& 337	& 257\\
& conversational entities & 26 & 	23\\
& conversational agents	& 3247	& 2721\\
& embodied conversational agents	& 1550 & 	1225\\
& human-computer conversational systems	& 0	& 0\\
\midrule
\multirow{8}{*}{arXiv}	& chatbot	   & 132   & 131	   & \multirow{8}{*}{56}\\
& conversational modelling	& 43	& 43\\
& conversational system	& 48	& 46\\
& conversation system	& 48	& 46\\
& conversational entities & 2	& 2\\
& conversational agents	& 77	& 77\\
& embodied conversational agents	& 4	& 4\\
& human-computer conversational systems	& 0	& 0\\
\midrule
\multirow{8}{*}{Google Scholar}	& chatbot	   & 36,000   & 16,400	   & \multirow{8}{*}{201}\\
& conversational modelling	& 183	& 116\\
& conversational system	& 4,510	& 2460\\
& conversation system	& 2850	& 2,190\\
& conversational entities	& 162	& 127\\
& conversational agents	& 23,600	& 16900\\
& embodied conversational agents	& 9960	& 7510\\
& human-computer conversational systems	& 26	& 8\\
\midrule
\multirow{8}{*}{JSTOR}	& chatbot	   & 318   & 311	   & \multirow{8}{*}{1}\\
& conversational modelling	& 1291	& 537\\
& conversational system	& 1962	& 498\\
& conversation system	& 1962	& 498\\
& conversational entities	& 31	& 14\\
& conversational agents	& 310	& 204\\
& embodied conversational agents	& 88	& 68\\
& human-computer conversational systems	& 0	& 0\\
\bottomrule
\end{tabular}
\end{adjustwidth}
\end{table}

\subsubsection{Reviewing~Articles}
We have reviewed large numbers of articles in order to identify some key aspects of chatbots from the identified literature sources. We conducted content analysis on 316 articles. In~this step we only read articles' abstract and identified their main objectives. This step was followed by further reviewing the 316 selected articles in order to identify key aspects of chatbots that have been the focus of previous~studies.  

By reviewing the selected 316 articles we have identified the following key aspects of chatbots that the literature sources have covered: 

\begin{itemize}
\item Chatbots' History and Evolution: this aspect encompasses all papers that presented a detailed description of chatbots' evolution over time. This category is fundamental since it helped us understand the trends and technologies that ascended or discarded over time, indicating the evolution of the chatbot. It also helped us discover how and why chatbots emerged and how their applications and purposes changed over time. Section~\ref{sec:background} offers overview of our finding on chatbots history and evolution.
\item Chatbots' Implementation: this aspect includes papers that present examples of chatbots architectural design and implementation. This category allowed us to identify the commonly used algorithms for chatbots and the specific algorithms that are used for diverse types of chatbots based on the purpose of chatbot application. This also allowed to identify the industry standards in terms of chatbots' models and algorithms, as~well as their shortcomings and limitations. Detailed implementation approaches to chatbots are given in Section~\ref{subsec:implementation-approaches}.
\item Chatbots' Evaluation: For this aspect, some articles focused on the evaluation methods and metrics used for measuring chatbots performance. It was important to identify these papers in order to understand the way chatbots are evaluated and the evaluation metrics and methods used. We outline the various evaluation metrics in Section \ref{subsec:evaluation}.
\item Chatbots' Applications: this aspect encompasses all examples of chatbots applied to a specific domain, such as education, finance, customer support and psychology. Papers pertaining to this category helped us tie information from previous categories and get a better understanding of what models and what features are used for which applications in order to serve different purposes. We identify and offer overview on the application of chatbots in Section~\ref{subsec:chatbot-application}.
\item Dataset: this category was used to classify chatbots depending on the dataset used to train machine learning algorithms for the development of language model. Section~\ref{subsec:databaset} highlights the main datasets that have been used in previous studies.
\end{itemize}

These categories emerged from the literature review. The~articles we reviewed covered one or more of these categories, often stated in the title or the abstract. It could be argued that these categories are widespread in the literature because they are strictly connected with chatbots development. Chatbot’s application development requires, in~fact, a~study of different implementations and other applications (which are the result of chatbots’ evolution over time), as~well as a dataset and an evaluation model. Any other aspect can be classified as a subcategory of one of these main~categories.

\section{Literature Review~Analysis}
\label{sec:literature}
In this section we outline the following  main aspects of chatbots based on our finding from the literature review: implementation approaches, available public database used in previous data-driven approaches to chatbot implementation, the~main evaluation methods for measuring the performance of chatbots and the application of chatbots in different~domains.

\subsection{Implementation Approaches to~Chatbots}
\label{subsec:implementation-approaches}
In this section, we will give an overview of chatbots' implementation methods. We will distinguish between Rule-based chatbots, and~Artificial Intelligence (AI) based chatbots. Within~AI-based chatbots, we will further distinguish among Information-Retrieval chatbots and Generative Chatbots. We will also discuss drawbacks and limitations of each implementation approach, as~well as recent~improvements.

\subsubsection{Rule-Based~Chatbots}
\label{subsubsec:rulebased-chatbot}
The very first attempts at chatbots' implementation were rule-based. Rule-based models are usually easier to design and to implement, but~are limited in terms of capabilities, since they have difficulties answering complex queries. Rule-based chatbots answer users' queries by looking for patterns matches; hence, they are likely to produce inaccurate answers when they come across a sentence that does not contain any known pattern. Furthermore, manually encoding pattern matching rules can be difficult and time consuming.Furthermore, pattern matching rules are brittle, highly domain specific, and~do not transfer well from one problem to the~other.

\subsubsection{Artificial Intelligence~Chatbots}
\label{subsubsec:ai-chatbots}
AI models, contrary to Rule-based models, are based on Machine Learning algorithms that allow them to learn from an existing database of human conversations. In~order to do so, they need to be trained through Machine Learning algorithms that can train the model using a training dataset. Through the use of Machine Learning algorithms, there is no longer the need to manually define and code new pattern matching rules, which allows chatbots to be more flexible and no longer dependent on domain specific knowledge. As~stated, AI models can be further categorised into Information Retrieval based models and Generative~models.

Information 
 Retrieval Models. Information Retrieval based models are designed so that given a dataset of textual information, the~algorithm will be capable of retrieving the information needed based on the user's input. The~algorithm used is usually a Shallow Learning algorithm; nonetheless, there are also cases of Information Retrieval models that use Rule-based algorithms and Deep Learning ones. Information Retrieval based models include a pre-defined set of possible answers; the chatbot processes the user query and based on this input it picks one of the answers available in its set. The~knowledge base for this kind of model is usually formed by a database of question-answer pairs. A~chat index is constructed from this database, in~order to list all the possible answers based on the message that prompted them. When the user provides the chatbot with an input, the~chatbot treats that input as a query, and~an Information Retrieval model akin to those used for web queries is used to match the user's input to similar ones in the chat index. The~output returned to the user is thus the answer paired with the selected question among those present in the chat index~\cite{shum_eliza_2018}. The~main advantage of this model is that it ensures the quality of the responses since they are not automatically generated. These models have seen a surge in popularity with the advent of the Web 2.0 and the increase in available textual information that could be retrieved on social media platforms, forums, and~chats~\cite{yan_learning_2016}. 

One of the main downsides of this approach is that creating the necessary knowledge base can be costly, time-consuming, and~tedious. Furthermore, if~the great volume of data available provides for a greater training set and a wider knowledge base, it also implies it will be all the more challenging to match a user's input to the correct answer; a significant amount of time and resources must be deployed to train the system to select one of the correct answers available~\cite{yan_learning_2016}. 

Finally, Information Retrieval systems, due to the fact that they do not generate answers but rather retrieve answers from a pre-defined set in their knowledge base, are arguably less suitable to be used as the underlying algorithm for conversational or chit-chat agents-the so-called social chatbots. Information Retrieval models are in fact less suitable to develop a personality, which is an important trait for this kind of chatbot~\cite{shum_eliza_2018}. 
Nonetheless, some progress has been made in developing new Information Retrieval algorithms in recent time, and~it is worth mentioning what Machine Learning algorithms are currently being used as underlying technology for this kind of model.~\cite{lu_deep_2013} proposed a new model to represent local textual co-occurrence and map hierarchical information across domains for more semantically distant terms. This model was based on the idea that the higher the co-occurrence of two terms across domains, the~more closely related the two terms are. Accordingly, a~high co-occurrence within a specific domain could inform the information retrieval process. This model was, thus, based on two steps: topic modelling for parallel text, and~getting the hierarchy architecture. The~first step aims at finding meaningful co-occurrence patterns of words. The~second step aims at modelling the architecture of co-occurrences across topics. This architecture will be used to create the neural network that powers this machine learning algorithm. The~interesting development made by this model lies therefore in its use of co-occurrences of words to define a context. The~underlying aim of this research was to use contextual information to improve matching performances for Information Retrieval models~\cite{lu_deep_2013}. 

One interesting development, which aims at taking into consideration previous turn in the conversation, thus obtaining more contextual information in order to improve the quality and the correctness of the output is the one proposed by~\cite{yan_learning_2016}. In~this model the Information Retrieval process is enhanced by a Deep Neural Network that ranks not only the question/answer pair matched with the last user's input, but~also those question/answer pairs that match with reformulated versions of previous conversation turns. The~ranking lists corresponding to different reformulations are then merged. In~this way, contextual information can be leveraged from the user's previous queries, and~these pieces of information can be used to retrieve a better answer within the knowledge base~\cite{yan_learning_2016}.

Generative 
 Models. Generative based models, as~the name suggests, generate new responses word by word, based on the input of the user. These models are thus able to create entirely new sentences to respond to users' queries; however, they need to be trained in order to learn sentence structure and syntax, and~the outputs can somewhat lack in quality or consistency~\cite{shang_neural_2015,sojasingarayar_seq2seq_2020,sordoni_neural_2015,vinyals_neural_2015,sutskever_sequence_2014}. 

Generative models are usually trained on a large dataset of natural phrases issued from a conversation. The~model learns sentence structure, syntax, and~vocabulary through the data that it has been fed. The~overall aim is for the algorithm to be able to generate an appropriate, linguistically correct response based on the input sentence. This approach is usually based on a Deep Learning Algorithm composed of an Encoder-Decoder Neural Network model with Long-Short-Term-Memory mechanisms to counterbalance the vanishing gradient effect present in vanilla Recurrent Neural Networks~\cite{vinyals_neural_2015}.

Industry-Standard 
 Algorithms. Among AI models, Sequence to Sequence models have become the industry standard for chatbot modelling. They were first introduced to solve Machine Translation problems, but~the underlying principles do in fact seem to perform well for Natural Language Generation as well. These models are composed of two Recurrent Neural Networks (RNN), an~Encoder and a Decoder. The~input sentence of the chatbot user becomes the input of the Encoder, which processes one word at a time in a specific hidden state of the RNN. The~final state represents the intention of the sequence and is called the context vector. The~Decoder takes the context vector as its input and generates another sequence (or sentence) one word at a time. The~overall objective for this probabilistic model is to learn to generate the most probable answer given the conversational context, which in this case is constituted by the previous turn in the conversation, or~the input sentence. In~the learning phase, the~answer, or~output sentence, is given to the model so that it can learn through back propagation. For~the interference phase, two different approaches can be used. The~beam search approach provides several candidates as the input sentence and the output sentence is selected based on the highest probability. A~greedier approach uses the predicted output token as an input to predict the next sentence in the conversation~\cite{vinyals_neural_2015}. 

This model does offer some interesting advantages. First, it does not involve domain-specific knowledge, but~is rather an end-to-end solution that can be trained using different datasets, thus on different domains. Furthermore, although~the model does not need domain-specific knowledge to provide valuable results, it can be adapted to work with other algorithms if further analysis on domain-specific knowledge is needed. It is thus a simple yet widely general and flexible model that can be used to solve different NLP tasks~\cite{vinyals_neural_2015,shum_eliza_2018}. For~these reasons, the~Sequence-to-Sequence model seems to have become the industry standard choice for dialogue generation and many NLP tasks in recent years. Nonetheless, it has a considerable limit: the entirety of the information contained in the input sentence must be encoded in a fixed length vector, the~context vector, and~thus, the~longer the sentence, the~more information gets lost in the process. That is why Sequence to Sequence models do not perform well when they must respond to longer sentences and tend to give vague answers. Furthermore, when generating an answer, these models tend to focus on a single response, which creates a lack of coherence in the turns of a conversation~\cite{sojasingarayar_seq2seq_2020,jurafsky_speech_2020,striger_end--end_2017}.

Transformers. 
Of course, one of the most interesting innovations in Deep Learning language models has been the introduction of Transformers, first presented by~\citep{vaswani_attention_2017}. in the paper “Attention is all you need”. Transformers are language models based solely on the Attention mechanism. Transformers are nowadays the model of choice for NLP challenges, replacing RNN models like long short-term memory (LSTM) by differentially weighing the relevance of each portion of the input data. Furthermore, they provide training parallelization that permits training on larger datasets than was originally achievable. This led to the development of pretrained systems such as BERT (Bidirectional Encoder Representations from transformers)~\cite{devlin_bert_2019} and GPT (Generative Pre-trained Transformer), which were trained with huge language datasets, such as Wikipedia Corpus and Common Crawl, and~may be fine-tuned for specific applications. Several different versions of the Transformer have since been presented, such as the Reformer~\cite{kitaev_reformer_2020} and the Transformer XL~\cite{dai_transformer-xl_2019}. Each version of the transformer has been developed to answer to specific challenges for the task at hand. Even though transformers were introduced to answer Machine Translation challenges, they can be adapted and modified to perform dialogue Modelling~tasks.

In ~\citep{dai_transformer-xl_2019}, the~authors propose an updated version of the Transformer called Transformer-XL. This model can go beyond the fixed-length context limitations of the Transformer, using sentence-level recurrence. Transformers show a potential of learning longer-term dependency but are constrained by fixed length context in the setting of language modelling. The~authors present a unique neural architecture named Transformer-XL that enables learning dependency beyond a given length without breaking temporal coherence. It comprises a segment level recurrence mechanism and a unique positional encoding technique. This solution aims at capturing longer-term dependency and resolving the context fragmentation issue. Even though this approach has not yet been applied to dialogue modelling, it can be argued that once the appropriate and necessary modification implemented, it could prove useful in overcoming some of the issues current dialogue models present, namely context~understanding.

In ~\citep{kitaev_reformer_2020}, the~authors introduce the Reformer, a~more efficient version of the Transformer, that makes use of two techniques to improve the Transformer in terms of efficiency. Firstly, the~authors substitute dot-product attention with one that employs locality-sensitive hashing, increasing its complexity from $O (L^2) to O (L log L)$, where L is the length of the sequence. Secondly, they employ reversible residual layers instead of the standard residuals, which permits storing activation only once in the training process instead of N times, where N is the number of layers. As~a result, the~Reformer is significantly more memory-efficient and substantially faster on longer~sequences.

In ~\citep{adiwardana_towards_2020}, the~authors introduce Meena, a~generative chatbot model that was trained end-to-end on 40 billion words extracted and filtered from public domain social media discussions. With~Meena, the~authors stretch the limits of the end-to-end approach in order to show that a big scale low-perplexity model can produce quality language outputs. The~authors employ a seq2seq model~\cite{sutskever_sequence_2014,bahdanau_neural_2016} with the Evolved Transformer~\cite{so_evolved_2019} as the main architecture. The~four most important characteristics of the Evolved Transformer's architecture are the utilization of (i) large depth-wise separable convolutions, (ii) Gated Linear Units~\cite{dauphin2017language}, (iii) branching structures and (iv) swish activations~\cite{ramachandran2017searching}. Both the Evolved Transformer's encoder and decoder independently generated a branched lower portion with wide convolutions. Also in both situations, the~later portion is essentially identical to the Transformer~\cite{so_evolved_2019}. The~model is trained on multi-turn dialogues where the input sentence is comprised of all turns of the context (up to 7) and the output sentence is the response. To~quantify the quality of Meena and compare it with other chatbots,~\citep{adiwardana_towards_2020} also propose an effective human evaluation metric. Sensibleness and Specificity Average (SSA) incorporates two basic qualities of a human-like chatbot: making sense and being specific in their response. The~authors ask human judges to evaluate responses based on these two~aspects.

\subsection{Datasets~Used}
\label{subsec:databaset}
In this section we will discuss the datasets that appear to be most frequently used to train deep learning chatbot models. Firstly, we must differentiate between open domain datasets, and~closed-domain datasets. Within~the literature, there appear to be a few open domain datasets that are most commonly used:

OpenSubtitles, Cornell, and~the DailyDialog dataset. We will discuss each one of these datasets more in detail. On~the other hand, there appear to be no closed domain dataset repeatedly used in the literature surveyed. This can be due to the specificity of such datasets, that usually attend to specific needs and objectives within a precise scope. For~brevity, we will not describe each one of these datasets in~details.

OpenSubtitles. 
The OpenSubtitles database is used by several works of literature. It is an open domain database made of movies' subtitles in XML format. The~database offers open access movies subtitles for several languages. The~dataset appears to be quite large, containing millions of sentence pairs, however the data does not appear to be of the best quality; scene descriptions, close captioning, and~segmented sentences appear among the dialogues, which can be problematic when training an open domain chatbot, since the cohesiveness of the dialogue can be lost~\cite{opensubtitles2016,christensen_context-aware_2018,zhong_affect-rich_2018,klein_opennmt_2017,li_deep_2016,li_dailydialog_2017,li_adversarial_2017,vinyals_neural_2015}.

Cornell.
The Cornell Movie-dialogue is also quite used to train open domain dialogue system. This corpus presents a metadata-rich collection of fictional conversations taken from raw movie screenplays, consisting of more than 300,000 total utterances. The~data is presented in `txt' format. Since the data has been taken from movie scripts, it appears to be of decent quality. However, the~dataset might not be large enough to train more advanced language models~\cite{Danescu-Niculescu-Mizil+Lee:11a,he_analyzing_2021,roller_recipes_2020,ghandeharioun_approximating_2019,zhong_affect-rich_2018} (No information was provided by the authors regarding the number of tokens in the dataset. For~this reason, we tokenized the entire dataset using the keras.preprocessing.text tokenizer object. Results are shown in Table~\ref{table:dataset_summary}.).

DailyDialog.
This dataset is human-made and consist of several multi-turn conversations. There are little over 13,000 conversations, each composed of 8 turns per speaker on average. The~data is annotated: annotations contain information regarding emotions and intents. These pieces of information can be useful in training some language models. Although~the data appears to be clean and less noisy compared to other datasets, it is also smaller, therefore less suited for more complex dialogue models~\cite{he_analyzing_2021,kim_multi-turn_2019,zhong_affect-rich_2018,li_dailydialog_2017} (No information was provided by the authors regarding the number of tokens in the dataset. For~this reason, we tokenized the entire dataset using the keras.preprocessing.text tokenizer object. Results are shown in Table~\ref{table:dataset_summary}.).

\begin{table}[H] 
\small
\caption{Summary of chatbot~dataset\label{table:dataset_summary}.}
\begin{tabular}{m{20mm} m{39mm} m{30mm} m{15mm} m{13mm}}
\toprule
\textbf{Dataset}	& \textbf{Content Type and Source} & \textbf{\# Phrases} & \textbf{\# Tokens} & \textbf{Source} \\
\midrule
OpenSubtitles &	Movie subtitles. Entire database of the OpenSubtitles.org repository &	441.5 M (2018 release) &	3.2 G (2018 release) &	\cite{opensubtitles2016}  \\
\midrule

Cornell	& Raw movie scripts. Fictional conversations extracted from raw movie scripts &	304,713	& 48,177 &	\cite{Danescu-Niculescu-Mizil+Lee:11a} \\

\midrule

DailyDialog	& Dialogues for English learners. Raw data crawled from various websites that provide content for English learners &	103,632 
(13,118 dialogues with 7.9 turns each on average) &	17,812 & \cite{li_dailydialog_2017} \\
\bottomrule
\end{tabular}
\end{table}
\unskip

\subsection{Evaluation}
\label{subsec:evaluation}
Evaluating dialogue systems has proven to be a challenging task, since human conversation attends to different goals and functions. Depending on the aim of the chatbot, the~metrics used to evaluate the dialogue can change. A~personal assistant chatbot will be mostly evaluated based on the effectiveness of the interaction (did the chatbot complete the task the user asked? Was the exchange efficient?), whereas a companion chatbot will be evaluated on its ability to keep the conversation going and to engage users. There are two main ways to evaluate a chatbot: human evaluation and automated evaluation~metrics. 

Human evaluation consists of asking a group of participants to interact with the chatbot, and~then evaluate the different aspects of the interaction according to an evaluation frameworks or questionnaire. Participants will usually rate the different aspects of the interaction based on a scale that can be used to draw averages and measure the quality of the performance in terms of efficiency, effectiveness, and~users' satisfaction~\cite{radziwill_evaluating_2019}. Although~human evaluation allows to assess the quality of different aspects of the interaction, it is costly (since there is a need to allocate human resources for the evaluation), time consuming, not easily scalable, and~subject to bias (even when following an evaluation framework, different people can rate the same interaction differently). Nonetheless, human evaluation can take into consideration several aspects of the exchange and assess the conversation at different levels; moreover, the~evaluation framework can be adapted based on the main aim and functions of the chatbot or dialogue system. For~these reasons, human evaluation metrics are used in several pieces of literature analysed, such as~\citep{christensen_context-aware_2018,sordoni_neural_2015}. The~PARAdigm for DIalogue System Evaluation is one of the most extensively utilized frameworks for combining different levels of evaluation (PARADISE). Firstly, PARADISE evaluates subjective variables such as (i) system usability, (ii) clarity, (iii) naturalness, (iv) friendliness, (v) robustness to misunderstandings, and~(vi) willingness to use the system again. It accomplishes this by soliciting user feedback via the dissemination of questionnaires. Secondly, through optimizing task success and decreasing dialogue costs, PARADISE aims to objectively quantify bot efficacy~\citep{cahn_chatbot_2017,walker_paradise_1997,hung_towards_2009,kempson_clarification_2021,li_acute-eval_2019,lin_xpersona_2020,sedoc_chateval_2019} propose different frameworks for human evaluation of chatbots. However, since these frameworks are reliant on human evaluation and therefore not easily scalable.
 ~\citep{przegalinska_bot_2019} The authors argue that trust is at the heart of effective human-chatbot interaction and examine how trust as a meaningful category is being redefined with the introduction of deep learning-enabled chatbots. However, the~proposed evaluation metric does not seem to consider efficiency, cohesiveness, and~overall dialogue~quality.

Automated evaluation metrics are more efficient in terms of time and resources necessary to carry out the evaluation. Nonetheless, there still appears to be a lack of industry standards in terms of evaluation metrics applied, and~automated evaluation metrics seem to lack the ability to correctly assess the quality, efficiency and effectiveness of the conversation as a whole. However, given the fact that these metrics can be more easily used, they are still widely implemented to evaluate chatbots. The~evaluation metrics used to measure accuracy will be standard evaluation metrics used for Machine Translation and other Natural Language Processing tasks such as BLEU, METEOR and TER, as~they have been used by~\cite{gelbukh_textual_2018,sordoni_neural_2015}. Although~these evaluation metrics are considered to be more suitable for Machine Translation problems, they can still provide valuable information regarding the Textual Entailment of the chatbot output~\cite{gelbukh_textual_2018}.

The F-score, alternatively referred to as the F1-score, is a statistic that indicates how accurate are e a model is on a given dataset. It is used to assess binary classification systems that categorize example evaluation s as either `positive' or `negative'. The~F-score is a measure of the model's precision and recall; it is defined as the harmonic mean of the model's precision and recall. The~equation is presented in \ref{fscore_equation}. The~F-score is frequently used to assess information retrieval systems such as search engines, as~well as numerous types of machine learning models, most notably in natural language processing. The~F-score can be adjusted to prioritize precision above recall, or~vice~versa. The~F0.5- and F2-scores, as~well as the normal F1-score, are often used adjusted F-scores. The~standard F1-score is calculated as the harmonic mean of the precision and recall. The~F-score of a perfect model is 1~\cite{noauthor_f-score_2019}. This evaluation metrics has been applied in a few research papers to evaluate chatbots performances, such as in~\cite{xu_recipes_2020,cuayahuitl_ensemble-based_2019}.
\begin{equation}
\label{fscore_equation}
F_{1}=\frac{\text{Precision} \times \text{Recall}}{\text{Precision} + \text{Recall}}
\end{equation}

Perplexity, abbreviated as PP, is the test set's inverse probability normalized by the number of words. Perplexity, however, is not applicable to unnormalized language models (that is, models that are not real probability distributions that sum to 1), and~it is incomparable for language models with different vocabularies~\cite{chen_evaluation_1998}. This metric has however been used in some recent studies to evaluate chatbots' performances, namely in~\cite{dhyani_intelligent_2020,john_legalbot_2017,higashinaka_towards_2014}. The~perplexity $PP$ of a language model $P_M(next word w history h)$ on a test set $T$ is computed with Equation~(\ref{perplexity_equation})
\begin{equation}
\label{perplexity_equation}
    PP_T({P_M}) = \frac{1}{(\prod_{i=1}^t PM(w_i|w_1...w_{i-1}))^{\frac{1}{t}}}
\end{equation}

bilingual evaluation understudy (BLEU) is widely used to assess various NLP tasks, even though it was first implemented to measure machine translation outputs. The~BLEU metric assigns a value to a translation on a scale of 0 to 1, however it is typically expressed as a percentage. The~closer the translation is to 1, the~more closely it resembles a human translation. Simply said, the~BLEU metric counts the number of words that overlap in a translation when compared to a reference translation, giving sequential words a higher score (KantanMT-Cloud-based Machine Translation Platform). ~\citep{dhyani_intelligent_2020,gelbukh_textual_2018,vaswani_attention_2017,sordoni_neural_2015} are some of authors that used BLEU scores to evaluate chatbots and other NLP tasks. However, BLEU does present some issues. BLEU's fixed brevity penalty does not effectively compensate for the absence of recall. Furthermore, Higher order N-grams are employed in BLEU as a proxy for the degree of grammatical well-formedness of a translation. It is argued that an explicit measure of grammaticality (or word order) can better account for the Machine Translation metric's weighting of grammaticality and result in a stronger association with human judgements of translation quality. Finally, BLEU presents an inadequate explicit Word Matching between Translation and Reference; although N-gram counts do not require specific word-to-word matching, this can result in inaccurate ``match''s especially for common function terms.
To compute BLEU score, First, the~geometric average of the modified $n$-gram precisions $P_n$ is computed using $n$-grams up to length $N$ and positive weights $w_n$ summing to one. $c$ is the length of the candidate translation and $r$ is the reference corpus length. Then, the~brevity penalty $BP$ is computed with the Equation  (\ref{blue_penalty_equaiton})
\begin{equation}
    \label{blue_penalty_equaiton}
    BP =  \left\{ \begin{array}{lrr}
    1 & \mbox{if} & c>r \\
    e^{(1-r/c)} & \mbox{if} & c\leq r
    \end{array}\right.
\end{equation}

Then, BLEU is computed with the Equation (\ref{bleu_equation})
\begin{equation}
    \label{bleu_equation}
    BLEU = BP \cdot \exp \left( \sum_{n=1}^Nw_n \log P_n \right)
\end{equation}

Metric for Evaluation of Translation with Explicit Ordering (METEOR) was created expressly to overcome the aforementioned problems in BLEU. It scores translations based on explicit word-for-word matches between the translation and a reference translation. If~many reference translations are available, the~given translation is evaluated independently of each reference and the best score is reported. This is explored in greater detail in the following section. METEOR produces an alignment between two strings when given a pair of translations to compare (a system translation and a reference translation). Alignment is defined as a mapping between unigrams in which each unigram in one string corresponds to zero or one unigram in the other string and to no unigrams in the same string. Thus, a~single unigram in one string cannot translate to more than one unigram in the other string within a given alignment~\cite{agarwal_meteor_2008,banerjee_meteor_2005}.  It is used by~\citep{sordoni_neural_2015} and by~\citep{xu_show_2015} along with BLEU to evaluate a chatbot model and an Image Captioning model, respectively. 
To compute the METEOR score for a sentence it's translation, first, unigram precision $P$ and unigram recall $r$ are calculated as $P = m/t$ and $R = m/r$, respectively, where $m$ is the number of mapped unigrams found between the two strings (a sentence and its translation), $t$ is the total number of unigrams in the sentence translation and $r$ is the total number of unigrams in the reference sentence. Then, the~parameterized harmonic mean of $P$ and $R$ is computed by Equation~(\ref{meteor_equation})
\begin{equation}
    \label{meteor_equation}
    F_{mean} =  \frac{P \cdot R}{\alpha \cdot P + (1-\alpha) \cdot R}
\end{equation}

Translation Error Rate (TER) has been used less compared to other method for evaluating chatbots performance, but~it is widely used to evaluate textual entailment. TER is a machine translation evaluation statistic that is calculated automatically. It is determined by the edit distance. It calculates the mistake rate by calculating the number of revisions necessary to convert a machine-translated output sentence to a human-translated reference sentence. Thus, the~complement of this error rate is considered when computing the similarity score~\cite{dhyani_intelligent_2020,snover_study_2006}. The~formulae for computer $TER$ score is presentation in Equation~(\ref{ter_equation})
\begin{equation}
    \label{ter_equation}
    TER = \frac{number\;of\;edit}{average\;number\;of\;reference\;words}
\end{equation}

Nonetheless, all these n-gram based evaluation models appear to be less fit to evaluate dialogue systems compared to other NLP tasks, because~two responses may have no overlapping n-grams, but~they can be equally effective in responding to a particular message. For~this reason, some recent work~\citep{kannan_adversarial_2017} has been conducted to study the usage of an adversarial evaluation method to evaluate dialogue models. Considering the effectiveness of generative adversarial networks (GANs) for image generation, the~authors propose that one indicator of a model's quality is the ease with which its output can be distinguished from that of a person. To~begin, they take a fully trained production-scale conversation model deployed as part of the Smart Reply system (the ``generator'') and train a second RNN (the ``discriminator'') on the following task: given an incoming message and a response, it must predict whether the response was sampled from the generator or from a human. The~purpose of this section is to determine whether an adversarial arrangement is viable for evaluation. The~authors can demonstrate that a discriminator can successfully separate model output from human output more than 60\% of the time. Additionally, it appears to expose the system's two key flaws: an erroneous length distribution and a dependence on typical, basic responses such as ``Thank you.'' Nonetheless, considerable difficulties with actual application of this strategy persist. There is still no indication that a model with a lower discriminator accuracy is necessarily superior in human evaluation. However, the~approach seems interesting since it essentially reproduces a Turing test in an automated and scalable manner.~\citep{kannan_adversarial_2017,li_adversarial_2017} suggest employing adversarial training for open-domain dialogue generation, drawing inspiration from the Turing test: the system is trained to generate sequences that are indistinguishable from human-generated sentences. Along with adversarial training, they describe a model for adversarial evaluation that leverages success in deceiving an adversary as a criterion for evaluating dialogues while avoiding a number of potential hazards. Refs.~\citep{ghandeharioun_approximating_2019,kuksenok_evaluation_2019} propose different evaluation frameworks. Nonetheless, the~proposed frameworks appear to be inadequate on open-domain, generative chatbots, and~have not been thoroughly~tested.

\subsection{Applications of~Chatbots}
\label{subsec:chatbot-application}
Chatbots are applied in many different domains. As~far as Education and Research go, chatbots in this domain seem to be mostly Information Retrieval or AIML based. Little to no Deep Learning application have been used in these fields. The~choice seems justified by the fact that chatbots created for educational purposes are often aimed at providing specific information (such as class schedules) or educational material. Refs.~\cite{ebner_potential_2020,arifi_potentials_2019,palasundram_sequence_2019,nwankwo_interactive_2018,bala_chat-bot_2017,fei_using_2013,augello_approach_2012,berger_conception_nodate} all provide examples of chatbots applied to Education and Research. For~similar reasons as in the field of education, most HealthCare oriented chatbots are Information Retrieval based.  Refs.~\citep{ayanouz_smart_2020,athota_chatbot_2020}, provide different examples of chatbots applications in~HealthCare.

E-commerce oriented chatbots present different configurations, mostly Information Retrieval based configurations, but~with some Deep Learning algorithms also involved in the overall architecture. This is possibly because in e-commerce, chatbots are often used to provide customer support. Therefore, they must be able not only to provide information on the products' catalogue and purchasing experience, but~also conversing with the customer.~Refs.\cite{cui_superagent_2017,ikumoro_intention_2019,singh_chatbot_2018} provide different examples of chatbots applied to~e-commerce.

Other Information Retrieval based chatbots applications can be found in Training~\cite{casillo_chatbot_2020}, Information Technology~\cite{melo_exploring_2020} and Finance~\cite{okuda_ai-based_2018}, possibly for similar~reasons.

Sequence to Sequence chatbots with attention mechanisms have been used to provide Law and Human Resource services, respectively in~\cite{john_legalbot_2017,sheikh_generative_2019}. In~the first case the chatbot is Long Short-Term Memory (LSTM) based, while in the second case it is based on Bidirectional-Gated Recurrent Units. 
As shown, when applied to a specific domains chatbots tend to fall back to Information Retrieval systems and Rule-based systems, or~a combination of the two. Only a few examples of applications have used Machine Learning technologies. This can be due to several~factors:
\begin{itemize}
\item Machine Learning in general and Deep Learning in particular, require a large amount of training data; although training data is becoming increasingly available but finding a suitable dataset might still represent a challenge. Furthermore, data needs to be pre\-processed in order to be used and might often contain unwanted noise.
\item Training is costly in terms of infrastructure and human resources, and~time consuming.
\item Chatbots, when they are not used for social or companion chatbots, are usually applied to a specific domain, which means that they require domain-specific training data (e.g., products information and details, financial information, educational material, healthcare information). This type of data is often confidential due to its nature; they are not readily available in open access to train a Deep Learning engine. Furthermore, given the nature of the data needed and of the tasks the chatbot is required to carry out (e.g., access a customer's purchase history, or~give more information about a product feature), Information Retrieval might be the best solution for most use-case~applications. 
\end{itemize}

In conclusion, we find from the literature a clear divide in terms of chatbots' technologies and their application. We observe that deep Learning algorithms trained on large open domain datasets, are usually implemented as social or companion chatbots. Task oriented chatbots appear to be usually trained on smaller, domain specific, and~often confidential datasets, and~they are usually based on Information Retrieval or Rule-based approaches, or~a combination of~both. 

\section{Related~Works}
\label{sec:related-work}
Previous literature survey work on different aspects of chatbots have focused on the design and implementation, chatbot history and background, evaluation methods and the application of chatbots in specific domain.  Our work is similar to previous work where we outline the background of chatbot. However, our paper differs from previous literature surveys where we discuss advances in chatbot design and implementation and state of the major limitations and challenges.~Ref.~\citep{abdul-kader_survey_2015} compare design techniques drawn from nine selected papers. The~authors focus especially on Loebner's winning chatbots, and~compare models used to develop those chatbots to the models presented in the selected papers.~Ref.~\citep{ketakee_chatbots_2017} discuss areas where chatbots fall short and explore research areas that need attention. The~survey conducted by~\cite{rahman_programming_2017} focused on cloud-based chatbot technology, chatbot programming and present and future programming issues in chatbots. The~authors conclude that stability, scalability and flexibility are the most important issues for consideration in chatbot development.~Ref.~\citep{cahn_chatbot_2017} conducts a study of the literature on the design, architecture, and~algorithms used in chatbots.~Ref.~\cite{bernardini_chatbots_2018} conducted a systematic literature review and quantitative study related to chatbot. They concluded by expressing concerns regarding the amount of published material and emphasized the importance of interdisciplinarity.~Ref.~\cite{nuruzzaman_survey_2018} compare the functionality and technical requirements of the eleven most common chatbot application~systems. 

The study conducted by~\citep{adamopoulou_overview_2020} involved two analysis of the literature that discuss the history, technology and applications of chatbots. While tracing the historical progression from the generative idea to the present day, the~authors highlighted potential shortcomings at each point. Following the presentation of a comprehensive categorization scheme, the~authors discussed critical implementation technologies. Finally, they discussed the general architecture of modern chatbots and the primary platforms for their creation. The~authors concluded that further research is needed on existing chatbots platforms and ethical issues related to chatbots. The~study by~\cite{singh_chatbot_2018} aimed at resolving the critical issue of identifying suitable deep learning techniques. They offered an overview of numerous commonly used deep learning systems models for learning. Additionally, they provided overviews of the entire design process, tips for implementation, and~links to several tutorials, analysis summaries and community-developed open-source deep learning pipelines and pre-trained models. They hoped that this survey will aid in the acceleration of the adoption of deep learning across several scientific~domains.

\section{Discussion}
\label{sec:discussion}
Despite current advancements in the fields of Deep Learning and Natural Language Processing, chatbots' architectures still present a few shortcomings. First and foremost, the~different language models proposed as chatbots' architecture are still unable to correctly mimic human conversation due to incorrect approach to dialogue modelling. The~underlying problem is that this model tries to solve conversational problems with a next-step approach: given an input, it tries to predict the best fitting output. This is, however, not the reasoning behind human conversation, that does not simply advance one step at a time, but~rather by taking into consideration a series of previous steps, the~underlying context of the conversation, and~the information being shared among the participants~\cite{vinyals_neural_2015}. Human conversation is not a step-by-step process as it is modelled in chatbots' architectures, but~rather a continuous journey, an~ongoing back-and-forth, where each step is dependent from the previous ones, or~subsequent ones. This dependency constitutes the conversational context, and~even though some new models have attempted at capturing such context~\cite{adiwardana_towards_2020,dai_transformer-xl_2019,christensen_context-aware_2018}, progress still must be~made. 

Another quite important shortcoming in chatbots' architecture is the apparent lack of a learned AI model for Information Retrieval chatbots. These chatbots, as~evidenced by the literature review on chatbots' applications, are widely popular across industries (e.g., healthcare, e-commerce, customer services and education) because they are able to provide coherent responses to a specific topic (e.g., booking an appointment, product specifics, returning an item, and~finding learning materials), given they can find a similar answer in their knowledge base. Currently, it seems that all the learned models for Information Retrieval chatbots depend on the dataset used to train them, and~there is no flexible learned model that can be applied to different datasets. It appears, in~fact, that research is now focused more on large generative models rather than smaller, easily implemented and domain independent. Such a model would find various applications across industries. The~challenge in developing such models is the lack of an open access domain-specific linguistic data, as~well as the highly diverse nature of industries and industry-specific topics where such models would be applied~to.
 	
 In terms of applications, as~shown we highlighted in the literature review Section~4, there is still a gap to be filled between the models used by the industry and recent improvements in the field. The~large models that represent the latest improvement in the field, whether they concern language models in general or dialogue models, are not yet suitable to be easily applied by the industry, as~they require great computational power and exceptionally large training datasets. As~we have previously stated, chatbots are applied across different industries to meet specific need. In~order to meet their purpose, chatbots' models are trained on specific datasets, and~their applications rely on different, often complex frameworks, that include dialogue managers and/or decision trees, and~are either knowledge-based or rule-based. Nonetheless, given the widespread use of such models, it appears evident that tailor-made solutions that ensure qualitative and precise answers to specific customers' queries are preferred over larger models that require a vast amount of data and perform better in open-domain conversation, but~might not perform as well in closed-domain one. It is evident there is a divide between open-domain (research-oriented models) and closed-domain (industry-oriented applications). To~bridge this gap, smaller, flexible, less domain dependent models would be beneficial. However, developing similar models is challenging because open domain, generative chatbots tend to be less precise and efficient in their~answers.
 
Regarding chatbot's evaluation, there are at least two major limitations that have yet to be addressed. Firstly, there is no common framework for chatbot's evaluation. Although~some metrics are widely used for measuring chatbots performance there is no specific metric or set of metrics commonly referred to as the reference. The~lack of common frame of reference concerning chatbots' evaluation limits the correct testing and comparison of different models. This limitation might also be due to the second limitation that emerges from the literature: the lack of reliable, efficient automatic evaluation method. As~we have stated in Section~\ref{sec:literature}, all automatic evaluation methods present some shortcomings, and~none are able to fully evaluate important aspects of dialogue quality, such as sentence coherence, cohesiveness, and~textual entailment aspects of chatbots. For~this reason, many models rely on human evaluation, but~human evaluation is costly, time consuming, not easily scalable, subject to bias, and~presents a lack of coherence. Furthermore, not even human evaluation presents a common frame of reference across models. To~overcome these limitations, a~new, reliable automatic evaluation method should be proposed. Such method should offer qualitative estimate chatbots' conversational outputs based on correctness, coherence, and~cohesiveness of the dialogue across multiple turns. Such an evaluation method could represent a crucial step forward in improving chatbots' performance~overall.

This literature survey has revealed several gaps in chatbot research that need to be addressed. Firstly, although~many survey papers on chatbots present a detailed explanation of chatbots' technologies and implementations, recent surveys lack information on most recent advances in language models that might be applied to chatbots, such as Transformers, which we have provided an overview of these advanced module in this paper. But~more in-depth analysis of said models and their application to chatbots would be beneficial. Similarly, truly little information and analysis on datasets is provided. The~type and quality of the data used to train Deep Learning models is particularly important to determine the output and the accuracy of the model. This is particularly true in language models, since the model has to learn the language task based on the linguistic data available, and~interpreting linguistic data is not as simple as interpreting numerical data. For~these reasons, it is important to discuss and analyse the data used to train the different models. Such analysis is also important because it allows for a fairer comparison of different models and their~performances. 

In this paper, we have provided an analysis of the most common open-domain datasets. Another crucial aspect of chatbots' implementation is their evaluation. Nonetheless, evaluation is not analysed in-depth in recent papers, and~although we have presented a few relevant pieces of literature concerning evaluation, as~we discussed, these focus more on single frameworks of evaluation rather than comparing several evaluation metrics. We have provided such an analysis and compared different evaluation metrics and discussed further steps to take to improve chatbots' evaluation. Finally, a~vital aspect of chatbots is their application in different industries and to real-life scenarios. Even though some papers provide an overview of chatbots' application and some numbers regarding their increase in popularity on the web, there appears to be a lack of clarity in identifying industry or task-oriented chatbots' applications and social or companion chatbots. Furthermore, this distinction is not drawn when discussing different chatbot models and their application either. In~our analysis we have tried to clarify this aspect, but~there is need for deeper and refined analysis and more clarity in future works of literature~survey.

Modelling chatbots is an interesting task that combines Deep Learning and Natural Language Processing, and~whose applications have incredibly grown in the past few years. Although~the first chatbots appeared sixty years ago, the~field has kept expanding and it presents new and engaging challenges. Their implementation across several industries and as companions and assistants creates many opportunities and fascinating paths of research, as~shown by the conspicuous amount of literature published on the subject in recent~year.

\section{Conclusions}
\label{sec:conclusion}
In this paper we have provided a survey of relevant works of literature on the subject, and~we have analysed the state of the art in terms of language models, applications, datasets used, and~evaluation frameworks. We have also underlined current challenges and limitations, as~well as gaps in the literature. Despite technological advancements, AI chatbots are still unable to simulate human speech. This is due to a faulty approach to dialogue modeling and a lack of domain-specific data with open access. For~Information Retrieval chatbots, there is also a lack of a learnt AI model. A~model like this might be used in a variety of sectors. There is still a gap to be closed in terms of applications between industry models and current advancements in the sector. Large models necessitate a lot of computing power and a lot of training data. There is no universal framework for evaluating chatbots. Several models depend on human evaluation, yet human evaluation is expensive, time-consuming, difficult to scale, biased, and~lacks coherence. A~new, reliable automatic evaluation approach should be provided to overcome these restrictions. Furthermore, recent studies have revealed a scarcity of data on the most recent developments in language models that may be used to chatbots like Transformers. As~a result, it's critical to examine and analyze the data used to train the various models. This type of study provides for a more accurate comparison of different models and their results. In~fact, the~distinction between chatbots' applications and social or companion chatbots appears to be hazy. Chatbot modeling is a fascinating challenge that mixes Deep Learning and Natural Language Processing. Despite the fact that the first chatbots were created sixty years ago, the~area has continued to grow and provide new and exciting problems. To~bridge these gaps, smaller, flexible, less domain dependent models would be beneficial. Improved, scalable, and~flexible language models for industry specific applications, more human-like model architectures, and~improved evaluation frameworks would surely represent great steps forward in the~field.

\vspace{6pt} 

\authorcontributions{Data curation, G.C.; Investigation, G.C. and S.J.; Methodology, G.C. and S.J.; Supervision, S.J. and K.M.; Writing---original draft, G.C.; Writing---review \& editing, S.J. and K.M. All authors have read and agreed to the published version of the manuscript.}

\funding{ This research received no external~funding.}

\institutionalreview{Not applicable 
}

\informedconsent{Not applicable 
}

\dataavailability{Not applicable 
}

\conflictsofinterest{The authors declare no conflict of~interest.} 

\begin{adjustwidth}{-\extralength}{0cm}
\reftitle{References}




\end{adjustwidth}
\end{document}